\documentclass[letterpaper, 10 pt, conference]{ieeeconf}  

\IEEEoverridecommandlockouts                              

\usepackage{graphicx}
\usepackage{hyperref}
\usepackage{caption}
\usepackage{subcaption}
\usepackage{multirow}
\usepackage[T1]{fontenc}
\usepackage{array}
\usepackage{float}
\usepackage{textcomp}
\usepackage{amsmath}
\usepackage{amssymb}
\usepackage{stackrel}
\usepackage{lipsum}
\usepackage{algorithmic} 
\usepackage{gensymb}
\title{\LARGE \bf
Reconfigurable Design for Omni-adaptive Grasp Learning*
}
\author{Fang Wan$^{1,\#}$, Haokun Wang$^{2,\#}$, Jiyuan Wu$^{2}$, Yujia Liu$^{2}$, Sheng Ge$^{2}$, and Chaoyang Song$^{3,*}$
    \thanks{*This work was supported by Southern University of Science and Technology and AncoraSpring Inc.}
    \thanks{\#Equal contribution as co-first authors.}
\thanks{$^{1}$Fang Wan is with AncoraSpring, Inc. and is a Visiting Scholar at SUSTech Institute of Robotics, Southern University of Science and Technology, 
        Shenzhen, Guangdong 518055, China. 
        Email: {\tt\small sophie.fwan@gmail.com}}%
\thanks{$^{2}$Haokun Wang, Jiyuan Wu, Yujia Liu and Sheng Ge are with
the Department of Mechanical and Energy Engineering, Southern University of Science and Technology, 
        Shenzhen, Guangdong 518055, China. 
        Email: {\tt\small {wanghk, wujy3, liuyj, 11612122}@mail.sustech.edu.cn}}%
\thanks{$^{3}$Chaoyang Song is the corresponding author with the Department of Mechanical and Energy Engineering, Southern University of Science and Technology,
        Shenzhen, Guangdong 518055, China.
        Email: {\tt\small songcy@ieee.org}}%
}
\begin{document}
\maketitle
\thispagestyle{empty}
\pagestyle{empty}
\begin{abstract}
    The engineering design of robotic grippers presents an ample design space for optimization towards robust grasping. In this paper, we adopt the reconfigurable design of the robotic gripper using a novel soft finger structure with omni-directional adaptation, which generates a large number of possible gripper configurations by rearranging these fingers. Such reconfigurable design with these omni-adaptive fingers enables us to systematically investigate the optimal arrangement of the fingers towards robust grasping. Furthermore, we adopt a learning-based method as the baseline to benchmark the effectiveness of each design configuration. As a result, we found that a 3-finger and 4-finger radial configuration is the most effective one achieving an average 96\% grasp success rate on seen and novel objects selected from the YCB dataset. We also discussed the influence of the frictional surface on the finger to improve the grasp robustness.
\end{abstract}
\section{Introduction}
\label{sec:Introduction}
    The reconfigurable design adopts the concept of modularity during the engineering integration of various functional components based on its operating environment \cite{huang1998modularity, Song2015a}. Robotic gripper, or the end-effector in general, provides the critical interaction between the robot system and target object in a particular operating environment. With human hands as the iconic inspiration for engineering design, most industrial grippers adopt a different strategy of under-actuation for cost-effectiveness \cite{monkman2007robot}. On the other hand, industrial grippers are usually designed with fewer actuators than that of the fingers to achieve a suitable or maximum adaptation for different objects during grasping. While a human hand has typically five fingers with 4 degrees of freedoms on each finger (the thumb has 5 DOFs) \cite{elkoura2003handrix}, a fundamental research question arises as to how many fingers shall be integrated to achieve maximum adaptation for robust grasping.
    
    Object-centric generalization provides a powerful representation of the grasp learning task. Multi-finger robotics for object manipulation has been a challenging issue due to the exponential increase in dimensions in related kinematics and dynamics \cite{Murray1994}. Recent progress shows a growing trend in adopting learning-based methods to solve this problem \cite{zeng2017multi,mahler2019learning}. Existing integration of robotic grippers usually adopts several fingers with different arrangements \cite{monkman2007robot}, ranging from 2-finger arrangement for minimum points of contact, 3-finger for added robustness, to 4/5-finger for human-like dexterity or load distribution. However, it remains an open question on the optimal number of fingers and their arrangements for an enhanced and robust grasping performance. 
    
    Robot learning is at the intersection between machine learning and advanced robotics, which benefits from recent development in data, computing, and algorithms \cite{connell2012robot}. The adoption of a learning-based method alleviates the modeling details of the interacting dynamics while extrapolating from feature-rich data analytics for statistically oriented learning algorithms \cite{levine2018learning}. It has been recognized that the adoption of soft and compliant grippers provides an implicit representation of the object variation in object-centric grasp generalization of manipulation skills and task models \cite{homberg2015haptic}. A direct benefit is a systematic reduction in the dimension of the manipulation problem during learning and execution \cite{manti2015bioinspired}.
    
    \begin{figure}[tbp]
        \centering
        \textsf{\includegraphics[width=1\columnwidth]{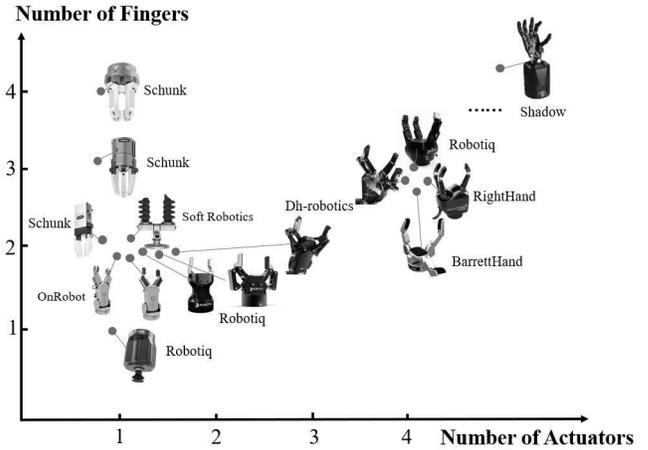}}
        \caption{Review of robotic grippers from multiple brands and models, and their finger arrangements expressed in terms of the number of fingers vs. actuators.}
        \label{fig:GripperReview}
    \end{figure}
    
\subsection{Related Work}
    Robotic fingers are generally inspired by animal fingers, especially those from the human, where delicate motor functions are supported by a comprehensive musculoskeletal system for dexterous manipulation \cite{schwarz1955anatomy}. While some research aims at an engineering replica of the human hand using robotics \cite{openai2018learning, xu2016design}, most industrial grippers are designed with a different strategy using just a few actuators and fingers due to cost-effectiveness, efficiency, and robustness in operation \cite{honarpardaz2017finger}. With the growing adoption of robotics in home automation, healthcare, logistics, etc., a growing need emerges for the robots to manipulate objects of a much larger variety than those commonly found at particular fixed work cells with only a handful of object variations \cite{singh2013evolution}. The need for robotic fingers with adaptive features becomes an original request in emerging research and applications \cite{hughes2016soft}. 
    
    With the bio-inspiration from humans, a 5-finger arrangement of the robotic fingers becomes a natural design choice. While it remains a research challenge to fully understand the biological reasons for five fingers per hand in most animals \cite{Raspopovic566}, some hints can be drawn from the constraint theory for object grasping \cite{Murray1994}. Parallel 2-finger gripper is the most common design that requires only one actuator to derive two fingers, usually arranged in parallel, for a minimum set of constraints for object grasping. Although, in theory, a third constraint is needed for a stable grasping, 2-finger grippers achieve stability through the frictional forces caused by the reaction force normal to the surface of object interaction \cite{lanni2009optimization}. More fingers can be added to the gripper to enhance the robustness of grasping by introducing more constraints \cite{butterfass2001dlr}. However, it remains a design superiority to utilize fewer actuators and fingers to achieve equivalent levels of dexterity, adaptation, and robustness when interacting with the physical environment.
    
    Soft robots further reinforce the concept of under-actuation in gripper design for enhanced adaptation. The nonlinear characteristics of the soft material under fluid or other forms of actuation produce a structural deformation across the whole body of the robot \cite{lee2017soft}, resulting in an unlimited number of DOFs for the benefit of adaptive motion \cite{hirose1978development}. The adoption of the learning-based method becomes a reasonable choice to deal with the adaptive grasping problem with implicit modeling of the motor functions and object variations \cite{choi2018learning}. However, there remains a research gap on the topological design optimization of the finger arrangement for robust grasping \cite{fan2018learning}, where the learning-based method may benefit from a systematic reduction of dimension in the hierarchical robot control.
    
\subsection{Proposed Method and Contributions}
    In this paper, we investigate the design optimization problem of finger arrangements in robotic grippers for enhanced grasping adaptation of daily-life objects using learning-based methods. A novel soft finger network with omni-directional, passive adaptation is adopted for the gripper design following a structural arrangement of the fingers for explorative experiments. By using a benchmark framework of DeepClaw for object manipulation, we collected 1000 blind grasps of YCB objects for training and evaluated 500 model-predicted grasps with different arrangements of the fingers. We found the statistical evidence that suggests a 3-finger and 4-finger radial arrangement of the soft finger networks achieve a grasp success rate of 96\% without the use of the rotational angle of the gripper, thus further reduced the control dimension without compromising grasp robustness. The contributions of this paper are listed as the following.
    \begin{itemize}
        \item The adoption of an omni-directional soft finger network with a friction enhanced layer for optimized grasping, where a reconfigurable gripper design is proposed for explorative study.
        \item A systematic experimentation of the finger arrangement for robust grasping, supporting the superiority of the radial 3-finger and 4-finger arrangements with 96\% grasp success rate.
        \item A dataset of blind grasping with various finger arrangement for reconfigurable gripper design, where a learning-based grasp planner is trained without the need for object-centric rotation during execution.
    \end{itemize}
    
    The rest of this paper is organized as follows. Section \ref{sec:Method} introduces the omni-adaptive finger network and the reconfigurable design of the gripper for finger arrangement research. Section \ref{sec:Results} explains the experimentation setup, procedure, and results, where a dataset is collected for training a learning model for universal grasping with different finger arrangements. Section \ref{sec:Discussion} discusses the results and summarizes the findings of this paper. Final remarks are included in section \ref{sec:Conclusion}, which conclude this paper.
    
\section{Reconfigurable Design for \\Omni-adaptive Grasping}
\label{sec:Method}
\subsection{Review of Finger Configurations in Gripper Design}
    We conducted a non-exhaustive survey of the standard multi-fingered grippers used in academic research and industrial applications, with results summarized in Fig. \ref{fig:GripperReview} and Table \ref{tab:GripperFeatures}. Unlike the Shadow hand that aims at replicating the motor functions of the human hand, most industrial grippers adopt either a radial configuration with all fingers facing the palm center or a parallel configuration with all fingers facing parallel to each other. The average numbers of fingers and actuators are either two or three. Some more advanced grippers with enhanced dexterity adopt four actuators or more. 
    
    \begin{table}[htbp]
        \caption{A non-exhaustive review of the design features in robotic grippers.}
        \label{tab:GripperFeatures}
        \centering
                \textsf{\includegraphics[width=1\columnwidth]{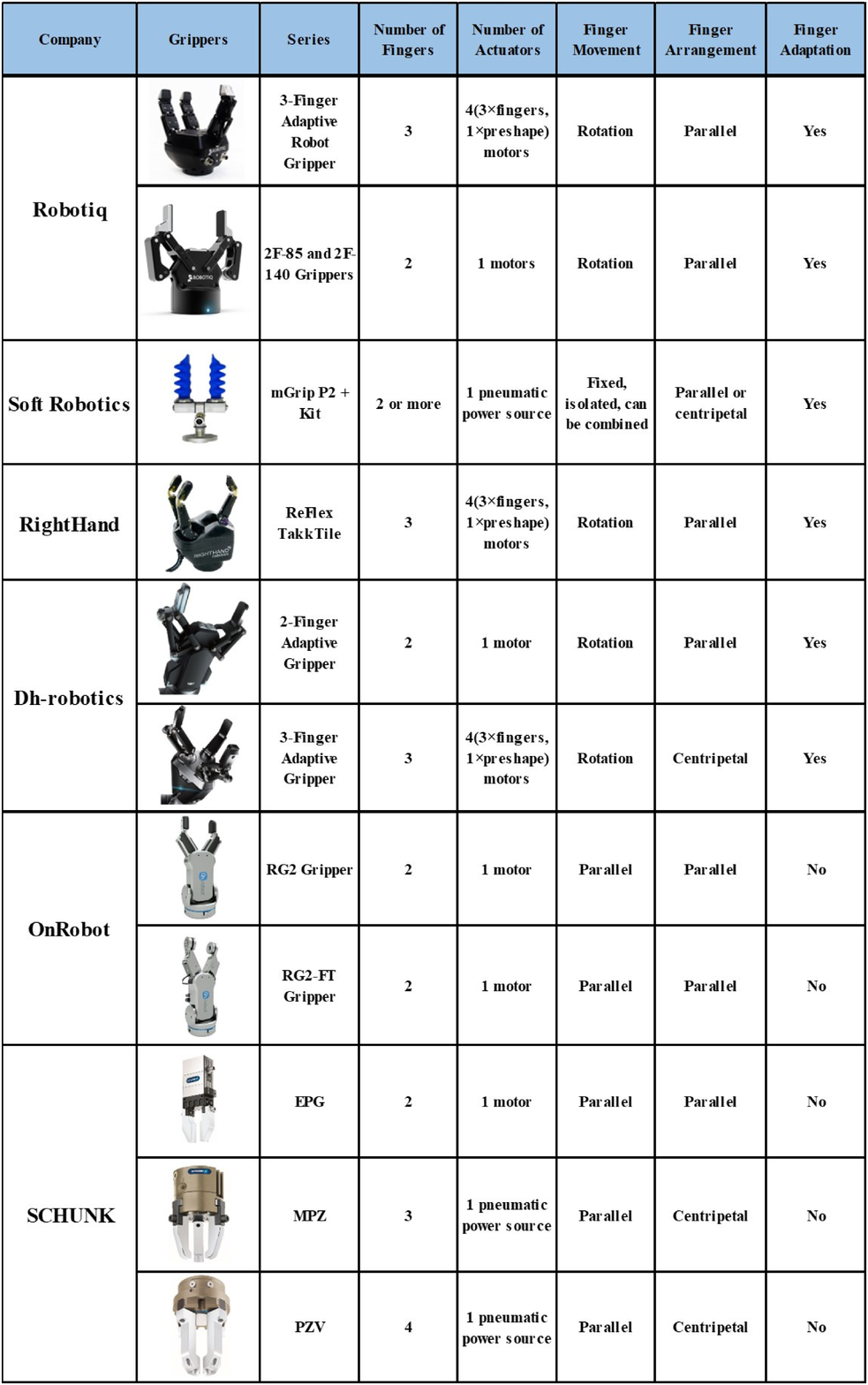}}
    \end{table}
    
    In general, one can observe a monotone increasing correlation between numbers of fingers and actuators against the price of the overall gripper. Under-actuation becomes an optimal design choice that overcomes the need for dexterity and reduces cost, where machine intelligence is introduced to achieve multi-modal operation in different operating conditions using fewer actuators than fingers. For example, the Robotiq's adaptive 3-finger and 2-finger grippers are classic examples of under-actuated grippers, where a preloaded spring is added to a five-bar mechanism to achieve transition between parallel mode and encompassing mode. A similar mechanism is also adopted by the DH-robot's grippers to achieve the same goal. The ReFlex gripper with three fingers from RightHand Inc. adopts four motors to drive the fingers, with three actuators driving fingers and one actuator for reconfiguration.
    
    Except for the under-actuation in mechanical robotic grippers, reconfigurable gripper design is also applicable to soft robotic grippers, such as those by the Soft Robotics Inc., where pneumatic driven fingers with patterned chambers of the cavity are arranged on the palm basis as a flexible gripper design. Due to the flexibility, light-weight, and low-cost of the silicone fingers, these pneumatic soft grippers are convenient for reconfiguration. The inflated soft fingers provide an infinite number of DOFs for under-actuated adaptation. 
    
    To sum up, the actuators and finger design are the two dominant factors in the gripper design. One should focus on the engineering specifications of the actuators and fingers when designing the gripper, which determines the gripper flexibility and function.
    
\subsection{Omni-directional Structural Adaptation}
\begin{figure}[htbp]
        \centering
            \textsf{\includegraphics[width=0.8\columnwidth]{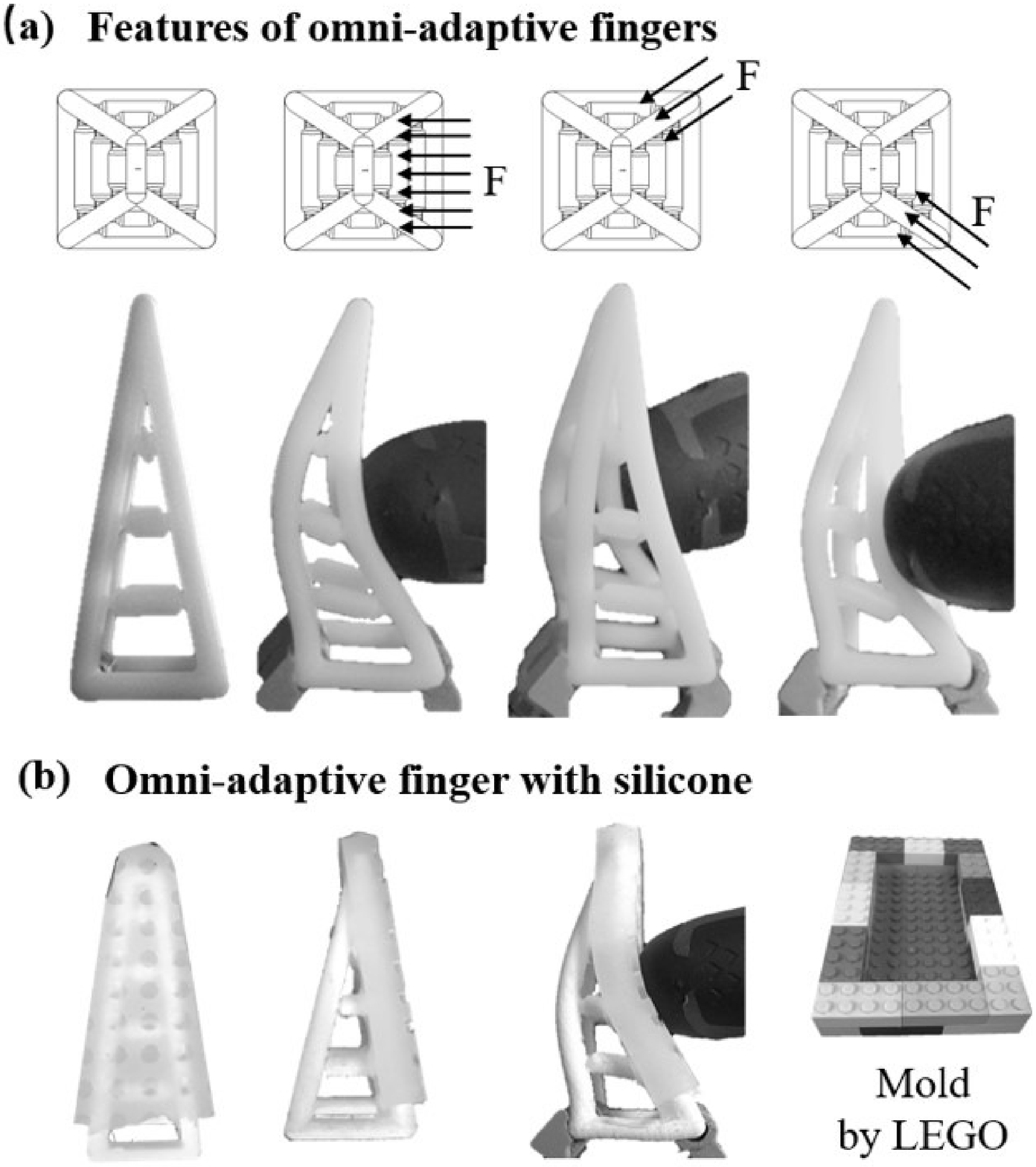}}
        \caption{The omni-directional adaptive design of the a soft finger network capable of passive adaptation in geometry during physical interaction in (a) and the finger surface design with a silicone rubber molded on the surface of grasping in (b), where a simple mold is fabricated using Lego.}
        \label{fig:SoftNetwork}
    \end{figure}

    The robotic grippers involve a large design space for engineering optimization, making it challenging to cross-compare even there are many commercial grippers on the market. In this paper, we propose to use a novel soft finger design shown in Fig. \ref{fig:SoftNetwork} with layered structure for omni-directional adaptation during physical interaction, which is cheap in cost, simple in design, safe during collision, flexible for integration, and scalable towards application. We designed a reconfigurable gripper structure where different number of these soft fingers can be easily rearranged in various configurations to explore the design space of finger arrangement in robotic grippers. 
    
    The proposed soft finger achieves omni-directional adaptation through a layered network structure with a converging form from the base towards the tip. Such asymmetric form enables a relatively large stiffness in the longitudinal direction, while the layer network structure enables a relatively low stiffness in the radial direction. On the other hand, one can easily modify the design by using different cross-sectional geometry from triangle to square, pentagon, hexagon, or even circular shape or any other geometry that suits the scenario. As a result, the differential stiffness design of the finger structure with the variable design of the cross-sectional geometry enables the soft finger to have omni-directional adaptation during physical interaction. Further analysis of the finger is beyond the scope of this paper, which will be addressed in another one.
    
\subsection{Design Reconfiguration for Robotic Grippers} 
    From the review, several design parameters can be summarized which contribute to the domain of design for robotic grippers, including the number of fingers $N$, the number of actuators $A$, and the geometric constraints of the finger arrangement $C$ in radial, parallel, or others. Let $G$ be the design system, a set of parameters specifying all options for configurations. An element $g \in G$ is defined as a configuration design. We give the function of $g$ as
    \begin{equation}
        \centering
            g=f(N,A,C)
        \label{eq:1}
    \end{equation}
    
    Since the omni-adaptive finger is passively deformed, only one air source is required to actuate the pneumatic cylinder under each finger base to control the opening and close of the gripper. Therefore, $N=A$ in our design. 
   
    \begin{equation}
        \centering
            g=f(N,C)
        \label{eq:2}
    \end{equation}
    So, 
    \begin{equation}
        \centering
            G={f(N_{1},C_1), f(N_2,C_2), f(N_3,C_3), ...}
        \label{eq:3}
    \end{equation}
    
\begin{figure}[htbp]
        \centering
            \textsf{\includegraphics[width=1\columnwidth]{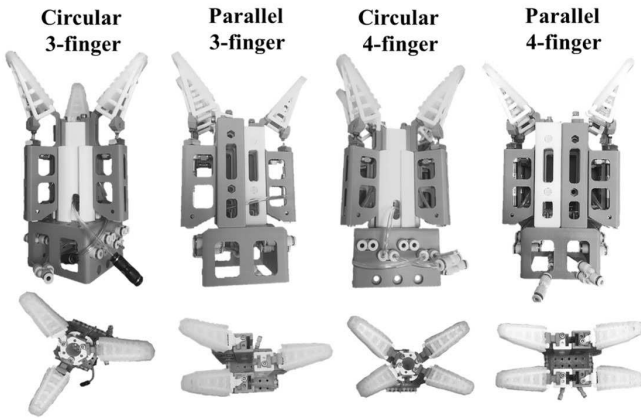}}
        \caption{Design reconfiguration of the gripper modules in radial or parallel with three or four of the omni-adaptive finger networks. All gripper configurations are actuated by one pneumatic input during operation.}
        \label{fig:ReconfigGripper}
    \end{figure}

    As shown in Table \ref{tab:GripperFeatures}, both parallel and radial arrangements of the fingers are commonly adopted in commercial gripper design. The parallel arrangement is generally easier in fabrication and assembly with the possibility of mounting many fingers opposite to each other in parallel arrangement, which is usually capable of picking up long size objects. The radial arrangement provides an even enclosure of the object from the radial directions for geometrical adaptation, but the number of fingers may be limited by the size of the gripper. As a result, we define $g$ to have five types of configurations, including 2-finger gripper, radial 3-finger (R3), parallel 3-finger (P3), radial 4-finger (R4), and parallel 4-finger (P4). In this paper, we focus our discussion on 3-finger and 4-finger grippers, which are less discussed in literature. The Fig \ref{fig:ReconfigGripper} shows the four types of reconfigurable grippers. 
    
    Each finger module is composed of an omni-adaptive finger, an air cylinder (SMC C8510-25,0.7MPa), and a mounting structure. Finger modules can be quickly reconfigured on the parallel palm or radial palm. Different combinations of palm and fingers result in the different gripper configurations, which changes the gripper features. Through the tracheal connection, the gripper can be controlled by the solenoid valve. The whole gripper system is compact and convenient for installation on the robot arm. 

\section{Experiment Results}
\label{sec:Results}
    The experiments are designed using robot learning methods to evaluate (1) the adaptability of the soft grippers; (2) the capability of the omni-adaptive soft gripper in reducing the dimensions of the grasp planning problem; (3) the grasp performance of different configurations of soft gripper.

\subsection{DeepClaw for Grasp Benchmarking}
\begin{figure}[htbp]
        \centering
            \textsf{\includegraphics[width=1\columnwidth]{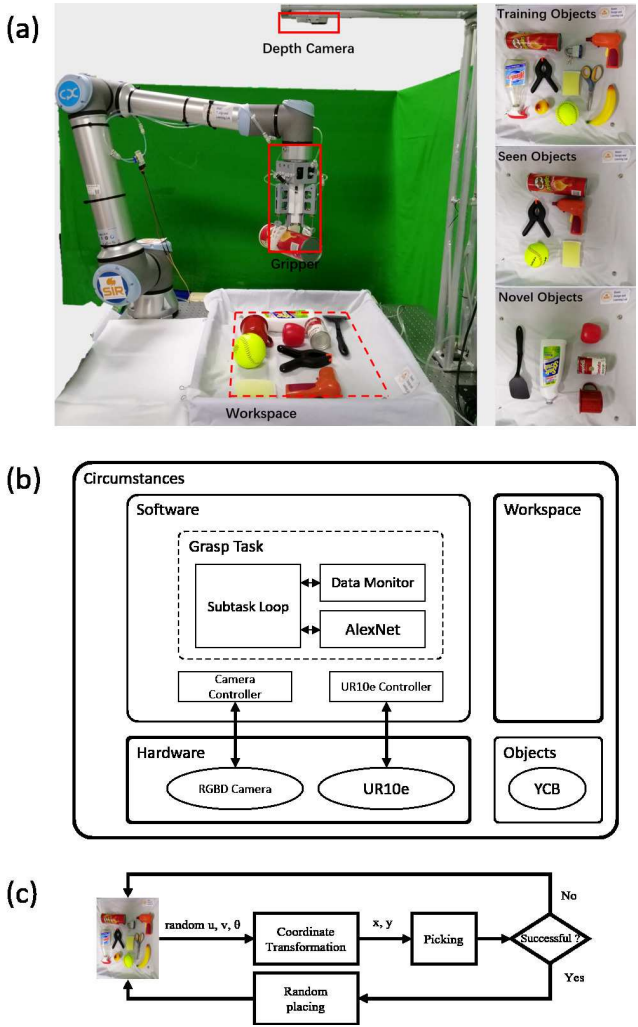}}
        \caption{(a) DeepClaw as an experiment setup for grasp benchmark. The system consists of a UR10e robot, a Realsense 435 depth camera, a designed soft gripper and a bin containing grasp objects. (b) The architecture of DeepClaw. (c) The collection process of training data.}
        \label{fig:ExpSetup}
    \end{figure}

    All the experiments ran on a desktop running Ubuntu 16.04 with four 2.5Ghz Intel Core i5 7300HQ and an NVIDIA 1050Ti. Physical grasping tests were run on DeepClaw benchmark system \cite{BionicDL2019}. The robot is a 6 DoF UR10e robot with the designed soft gripper mounted on the tool flange, as shown in Fig \ref{fig:ExpSetup}(a). The robot was mounted on a table, and a Realsense 435 depth sensor was mounted about 1 meter above the table, providing $1280\times720$ resolution color and depth images. A rectangle bin containing objects was placed underneath the depth sensor. The actual workspace is $40cm\times50cm$. The robot and the depth sensor were calibrated to obtain the hand-eye matrix $H$. The software and hardware architecture of DeepClaw is shown in Fig \ref{fig:ExpSetup}(b). The software contains driver modules for cameras and robot arms, subtask pipeline \cite{yang2020learning}, and data monitor for recording experiment data. A subtask pipeline consists of functionalities to locate and grasp the objects. In this work, we use an end-to-end learning method to detect objects and plan grasps.
    
    Soft grippers are well known to have excellent adaptability \cite{7988691,7989556} and are able to grasp a variety of objects while keeping perpendicular to the tabletop in applications \cite{chin2019automated}. Our hypothesis is that the grasp planning problem with our designed soft gripper can be simplified to predict the best grasp pose $(u,v,\theta)$ where $\theta$ is the yaw angle of the soft gripper while the gripper was kept perpendicular to the table. Due to the soft and adaptive nature of the fingers, grasps would be most stable when the object is fully embraced by the soft fingers as long as the object does not collide with the palm of the soft gripper. Hence the optimal grasp position in z-axis with reference to the sensor is defined by the following strategy.
    
    \begin{equation}
        z=
        \begin{cases}
            z_{bin}-\delta h, & H_{obj}<H_{finger}\\
            z_{obj}+H_{finger}-\delta h, & H_{obj}>H_{finger}
        \end{cases}
        \label{eq:ZStrategy}
    \end{equation}
    where $z_{bin}$ is the depth of the bottom of the bin, $z_{obj}$ is the depth at given $(u,v)$ read from the depth sensor, and $\delta h$ is a tiny offset to avoid collision with the bin. With known intrinsic parameters of the sensor and hand-eye transformation matrix, the grasp position $(u,v,z)$ was then transform to $(x,y,z)$ with reference to the robot arm for grasp execution.
    
\subsection{Learning Grasp Planners}
\label{subsec:DataTraining}
    The grasp trials were divided into two phases. Phase one collected 1000 grasp attempts using 3-finger radial soft gripper, which were later used to train grasp planners using the convolutional neural network (CNN). Ten objects from the YCB dataset were chosen with diverse sizes, geometries, and weights, as shown in Fig \ref{fig:ExpSetup}(a). 

    The collection process of training data is summarized in Fig \ref{fig:ExpSetup}(c). For each grasp attempt, five YCB objects were placed in the workspace, and the robot stood by at the home position. The Realsense 435 took a color and a depth image that captures both the objects and the soft gripper. Then the robot proceeded to perform a blind grasp with random $u$, $v$, and $\theta\in\left[-\frac{\pi}{2},\frac{\pi}{2}\right)$. Then the robot raised the object and transported it to the home position where the Realsense 435 took another color image and decided whether the grasp succeed or fail. A grasp was labeled as a success if it was able to lift and transport a desired object to the home position. The object was placed back into the bin, and the robot returned to the home position before a new grasp attempt started. Each entry of the training dataset contains a color image, and a grasp pose $(u,v,\theta)$, and a label indicating successful or failed grasps. There were 270 successful grasps among the 1000 attempts.
    
\begin{figure}[htbp]
        \centering
            \textsf{\includegraphics[width=1\columnwidth]{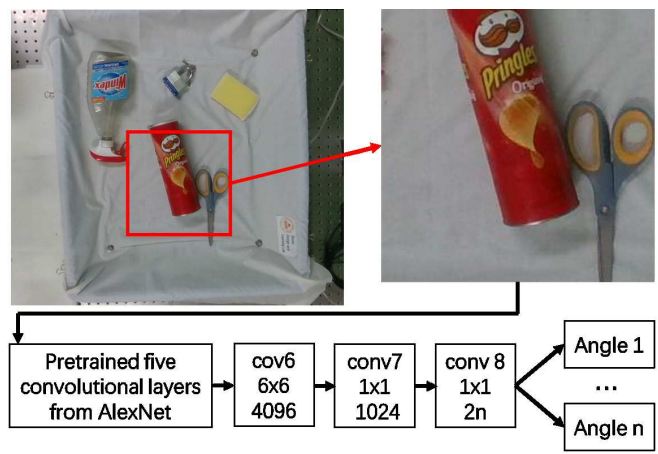}}
        \caption{The architecture of the grasp planner network adapted from AlexNet by converting the last three fully connected layers to $1\times1$ convolutional layers. The last layer output $n$ binary classifications. }
        \label{fig:GraspNetwork}
    \end{figure}

    In order to evaluate the capability of the omni-adaptive soft gripper in reducing the dimensions of the grasp planning problem, we trained and compared two CNN-based grasp planners on the training dataset with or without rotation angle information. We built a fully convolutional neural network (FCN) adapted from AlexNet \cite{krizhevsky2012imagenet}, as illustrated in Fig. \ref{fig:GraspNetwork}. During training time, the network takes as input a cropped color image centered at the grasp position $(u,v)$ such that the grasp position information is embedded in the image itself. The cropped patch size is $250\times250$, which covers the soft fingertips on the image and is resized to $227\times227$ before fed to the AlexNet convolutional layers. The network predicts the successful grasp probabilities independently for n rotation angles where $n$ is set to 1 or 9 in this work to test the capability of the omni-adaptive soft gripper in reducing the dimensions of the grasp planning problem. In the case of n equals 1, the network predicts a single successful grasp probability for position $(u,v)$ regardless of the rotation angle of the soft gripper. In the case of n equals to 9, the network predicts a 9-dimensional probability vector where each element represents the success probability of grasping position $(u,v)$ at -80\degree, -60\degree, \dots, 60\degree, 80\degree.

    The first five convolutional layers were initiated and fixed with weights pre-trained on ImageNet, and the last three layers were initialized with a truncated normal distribution. The loss function of the network is defined similarly to that in \cite{pinto2016supersizing} such that only the loss corresponding to the grasp angle is backpropagated. We used Tensorflow with a batch size of 128. The training accuracy converged to 1 quickly within 80 training steps when $n$ equals to 1 and 50 training steps for $n$ equals to 9. In the rest of this work, we shall call the trained network with $n$ equals 1 and 9 grasp-planner-1 and grasp-planner-9, respectively.
        
\subsection{Evaluating Design Reconfiguration}
    Phase two of the grasp trials was to evaluate the performance of the learned grasp-planner-1 and grasp-planner-9 on five known and five novel objects using different soft gripper configurations as detailed in Table \ref{tab:ExpProcedure}. The collection process is similar to that in Fig. \ref{fig:ExpSetup}(c) except the random $u$, $v$, and $\theta$ are replaced by values predicted by the learned grasp planners. We performed five sets of grasp trials, and each set consists of 100 trials (ten per object). Since our planners are fully convolutional neural networks, we can feed a color image of any size to the network and obtain a relatively dense probability map at a single prediction. The computation time for a single prediction was about 90 ms for both planners. 
    
    For each grasp attempt, one object was placed in the bin, and we fed the color image of the whole bin to the grasp planners and obtained the probability map of successful grasps. Then the grasp with the highest success probability was executed. Then the object was randomly placed back into the bin for the next grasp trial. The resolution of the map is determined by the size of the input image and the strides used in the network. With our architecture, the map gives a prediction every 32 pixels. When executing grasps predicted by grasp-planner-1, we fixed the orientation of the soft gripper. Since the objects were randomly placed in the bin, these grasps can be regarded as a random grasp in terms of $\theta$.
    
    \begin{table}[htbp]
        \caption{Grasping test experiments designed to evaluate four different finger configurations.}
        \label{tab:ExpProcedure}
        \begin{center}
            \begin{tabular}{|c|c|c|}
                \hline
                No. & With or w/o rotation & Finger configuration\\
                \hline
                1 & with & R3: 3-finger radial\\
                \hline
                2 & without & R3: 3-finger radial\\
                \hline
                3 & without & P3: 3-finger parallel\\
                \hline
                4 & without & R4: 4-finger radial\\
                \hline
                5 & without & P4: 4-finger parallel\\
                \hline
            \end{tabular}
        \end{center}
    \end{table}

\subsection{Results}
\begin{figure}[htbp]
        \centering
            \textsf{\includegraphics[width=1\columnwidth]{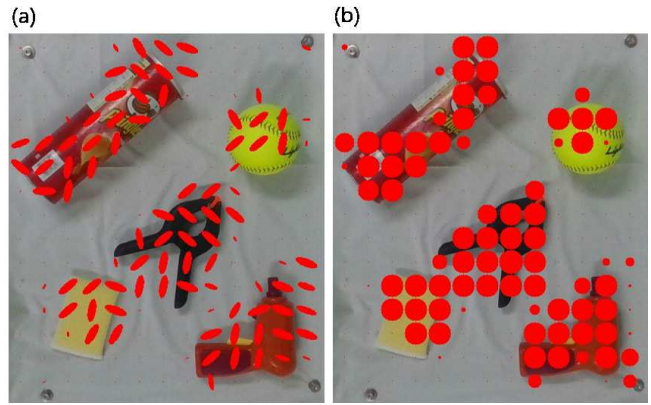}}
        \caption{The probability maps of successful grasps predicted by (a) grasp-planner-9 where the length and orientation of long axis of the oval represent the highest success probability and its corresponding grasp angle at the center of the oval; (b) grasp-planner-1.}
        \label{fig:Result-1-ProbMap}
    \end{figure}

    Fig. \ref{fig:Result-1-ProbMap} shows examples of probability maps predicted by grasp-planner-9 and grasp-planner-1. We found that both planners have successfully learned to detect the location of the objects from the background as they predicted high probabilities around the objects and low probabilities at empty areas. Besides, the two maps produced quite similar probability distributions, which led us to the hypothesis that with the exceptional adaptability of the soft fingers, it might be possible to achieve a high success grasp rate even without predicting the grasp angles.
    
\begin{figure}[htbp]
        \centering
            \textsf{\includegraphics[width=1\columnwidth]{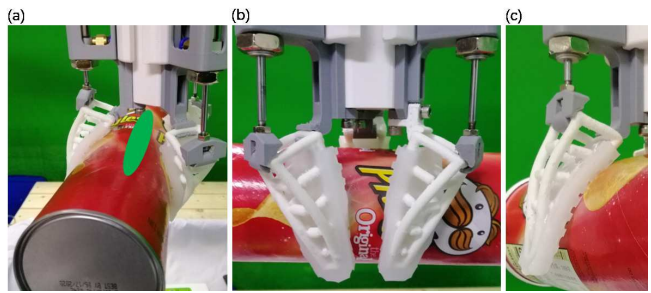}}
        \caption{(a) The deformation of the soft fingers when grasping the Pringles can, the green oval is the optimal grasp prediction; (b) two fingers side and (c) one finger side.}
        \label{fig:Result-2-AdaptFinger}
    \end{figure}

    Fig. \ref{fig:Result-2-AdaptFinger} shows the deformation of the soft fingers of the 3-finger radial gripper when grasping the Pringles can. All three fingers were able to adapt to the shape of can in different ways and form a firm grasp.

\begin{figure}[htbp]
        \centering
            \textsf{\includegraphics[width=1\columnwidth]{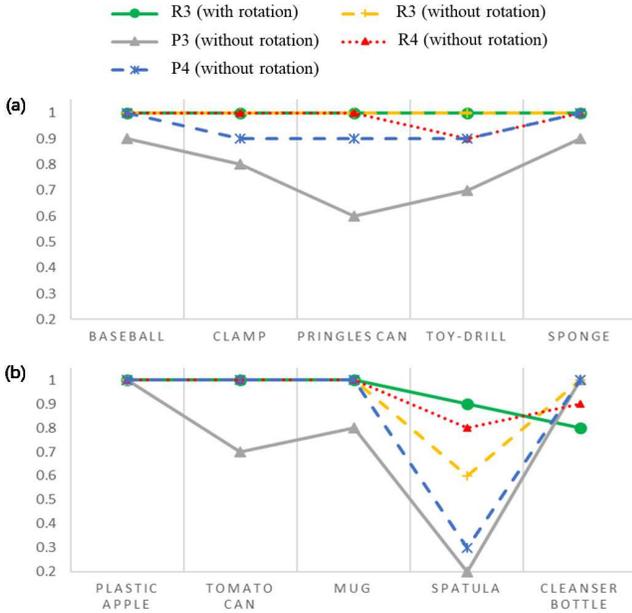}}
        \caption{Grasp success rates of (a) 5 seen objects and (b) 5 novel objects using four different finger configurations}
        \label{fig:Result-3-GraspRate}
    \end{figure}

    The grasp evaluation results are shown in Fig \ref{fig:Result-3-GraspRate}. The 3-finger radial soft gripper achieved comparable performance with grasp-planner-9 and grasp-planner-1 and was able to grasp all seen objects at $100\%$ success rate. Even for unseen objects, the 3-finger radial configuration achieved an average $94\%$ and $92\%$ success rate with grasp-planner-9 and grasp-planner-1, respectively. 
    
    Both 3-finger and 4-finger radial configurations achieved an average $96\%$ success rate over the ten testing objects with grasp-planner-1. 4-finger radial configuration outperformed 3-finger configuration when grasping the spatula but under-performed when grasping the cleanser bottle.
    
\section{Discussion}
\label{sec:Discussion}
\subsection{Optimized Finger Configuration for Adaptive Grippers}
    Our experiment results generally support an optimized gripper design with three or four fingers arranged in a radial configuration for robust grasping outcomes using the proposed soft finger networks. Considering the engineering trade-offs, we recommend the 3-finger radial arrangement for cost-effectiveness and space-saving with maximum usage scenarios. On the other hand, we recommend the 4-finger radial arrangement for enhanced robustness with higher and even payload distribution with redundant form closure for forceful grasping. Although the results presented in this paper are biased towards to soft finger network used in this paper, the qualitative outcome is generally applicable to grippers of other material properties. While 2-finger arrangement is typical among gripper of rigid design, we recommend grippers with soft fingers to follow our design recommendation of finger arrangement for a robust grasping outcome.
    
    It should be noted that the fingers presented in this paper adopts an angled output when closing the fingers, which naturally pushes the target object away towards the desktop or sideways, resulting in reduced stability in grasping outcomes. As shown in Fig. \ref{fig:GripperReview}, such angled grasping is also avoided by most industrial grippers except for the one developed by the Soft Robotics, Inc. due to its unique form closure during grasping. We expect the results to be further improved if we change the angled grasping to parallel grasping by changing the mechanism between the fingers and the actuators. 
    
    The results shown in Fig. \ref{fig:Result-3-GraspRate} also suggest that for grasp success rate does not change very much among the finger arrangements experimented in this paper. One exception is with the parallel 3-finger arrangement, which performs poorly for both known and new objects. By analyzing the failed grasps, it becomes evident that the encompassing grasping is not suitable for the soft finger network used in this paper. The side with two fingers has a gap of 1 finger width, which is not suitable to pick up objects of relatively small size. This explains the result in Fig. \ref{fig:Result-3-GraspRate}(b) in the solid grey line, which is even worse when the z-axis rotation is removed. Please note that for rigid grippers, such a 3-finger configuration with encompassing closure is an advantageous arrangement, i.e., Robotiq's adaptive 3-finger gripper. This is similar in the parallel 4-finger arrangement. Note that in this case, the grasp success rate for known objects are still very high, and for the five new objects, the grasp success rate achieved 100\% for four objects except spatula, which is of an irregular geometry and slim shape. 
    
\subsection{Dimensional Reduction for Grasp Learning}
    As shown in Fig. \ref{fig:Result-3-GraspRate}, our results indicated the potentials of removing the z-axis rotation for general-purpose object grasping. For the five known objects, the trained planner without rotation achieved 100\% grasp success for seen objects. For the five novel objects, the trained planner still achieved a 100\% grasp success rate for simple objects, including plastic apple, tomato can, mug, and cleanser bottle, but dropped slightly for the spatula. One should notice that the spatula is of a somewhat irregular shape and is also considered challenging to pick up in general.
    
    This potentially simplifies the grasp planning problem to a simple localization problem. In scenarios where grasp items are separated from each other, we could locate the object using computer vision or deep learning methods and perform the grasp at the centroid without the need to collect grasp training data.
    
\subsection{Friction Enhanced Finger Design for Robust Grasping}
    The silicone skin covering the soft fingers enhanced the frictions between the fingers and objects greatly. To test its effect, we also performed a set of 100 grasp trials (10 per object) similar to grasping test No.2 in Table \ref{tab:ExpProcedure} except the silicone skin was removed from the fingers. As shown in Fig, the average success rates of ten objects dropped considerably from $96\%$ to $34\%$. Among the ten objects, the success rates of sponge and tomato can remain relatively high at $90\%$, and the success rates of Pringles can, mug and spatula dropped to zero. Further investigation of the design of silicone skin will be explored in our future work.
    
    \begin{figure}[htbp]
        \centering
            \textsf{\includegraphics[width=1\columnwidth]{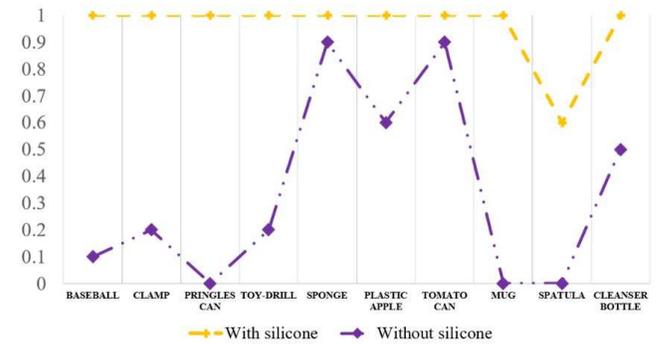}}
        \caption{Grasp success rates of 5 seen objects and 5 novel objects using R3 finger configuration with (yellow) and without (purple) silicone skin covering the soft fingers.}
        \label{fig:Result-4-GraspRate}
    \end{figure}

\subsection{Learning From Failures for Effective Model Training}
    Besides the training dataset described in section \ref{subsec:DataTraining}, we also tried to place all the ten training objects in the bin and collected 2000 bind grasp attempts, whose success rate was $41.4\%$. However, the learned grasp planners failed to learn the position information of the objects and give meaningful predictions. To investigate the possible reasons, we purposely selected 250 successful grasps, and 750 failed grasps from 2000 grasp to train the network. The learned grasp planner was very aggressive and tended to give high probabilities even at empty spaces. The learned lesson is that we have to leave enough space between objects in order to learn from failures.

\section{Conclusion}
\label{sec:Conclusion}
    In this paper, we explored the reconfigurable design for finger arrangement using a novel soft finger with omni-directional adaptation. We adopted the robot learning method to experiment with different arrangement of the fingers for design guidelines of a robust robotic gripper. In particular, our result shows that the 3-finger radial configuration is suitable for space-saving and cost-effectiveness, whereas the 4-finger radial arrangement can be applied to cases that require a higher payload with even distribution. We also achieved dimension reduction using the proposed gripper design with the removal of z-axis rotation during grasping. We also reported the different outcomes with or without friction enhancement of the soft finger network. Although our proposed gripper achieved a high success rate even during the blind grasping stage, we found that it is necessary to intentionally include enough failed grasps during the data collection stage to improve the trained model.
    
    The limitation of this work is the focus of design parameters on finger arrangement, FEM analysis of the soft finger, and the limited experiment with the angled grasping, which will be addressed in the future. Further testing is required to involve more objects to verify the design guidelines with statistical evidence.

\addtolength{\textheight}{-6cm}   

\bibliographystyle{IEEEtran}
\bibliography{IEEEexample}
\end{document}